\RequirePackage{fix-cm}
\documentclass[smallextended]{svjour3}       
\smartqed  
\usepackage{graphicx}
\usepackage{amssymb}
\usepackage{amsmath}
\usepackage[flushleft]{threeparttable}
\usepackage{cite}
\usepackage{bbm}
\usepackage{url} 
\usepackage{booktabs}
\usepackage{multirow} 
\begin{document}

\title{Prediction of Survival Outcomes under Clinical Presence Shift: A Joint Neural Network Architecture}

\titlerunning{Prediction of Survival Outcomes under Clinical Presence Shift}        

%
%

\author{Vincent Jeanselme         \and
	Glen Martin \and
	Matthew Sperrin \and
	Niels Peek \and\\
	Brian Tom \and
	Jessica Barrett
}


\institute{Vincent Jeanselme \at
	Department of Biomedical Informatics, Columbia University\\
    MRC Biostatistics Unit, University of Cambridge\\
	\email{vj2292@columbia.edu}           
    \and Glen Martin \and Matthew Sperrin \at
    Division of Informatics, Imaging \& Data Sciences,
    University of Manchester 
    \and Niels Peek \at
    The Healthcare Improvement Studies (THIS) Institute, 
    University of Cambridge
    \and Brian Tom \and Jessica Barrett \at
    MRC Biostatistics Unit, 
    University of Cambridge
}

\authorrunning{Jeanselme V. et al.} 

\date{}

\maketitle

\begin{abstract}
Electronic health records arise from the complex interaction between patients and the healthcare system. 
This observation process of interactions, referred to as \emph{clinical presence}, often impacts observed outcomes. 
When using electronic health records to develop clinical prediction models, it is standard practice to overlook clinical presence, impacting performance and limiting the transportability of models when this interaction evolves.
We propose a multi-task recurrent neural network that jointly models the inter-observation time and the missingness processes characterising this interaction in parallel to the survival outcome of interest. 
Our work formalises the concept of clinical presence shift when the prediction model is deployed in new settings (e.g. different hospitals, regions or countries), and we theoretically justify why the proposed joint modelling can improve transportability under changes in clinical presence. 
We demonstrate, in a real-world mortality prediction task in the MIMIC-III dataset, how the proposed strategy improves performance and transportability compared to state-of-the-art prediction models that do not incorporate the observation process. 
These results emphasise the importance of leveraging clinical presence to improve performance and create more transportable clinical prediction models. 
\end{abstract}

\section{Introduction}
\label{intro}

Due to their availability and cost-saving potential, Electronic Health Records (EHRs) are often considered a de facto resource for developing clinical prediction models~\cite{goldstein2017opportunities}. Unlike controlled experiments, EHRs reflect observational data arising from real-world interactions between patients and healthcare providers. \emph{These interactions are not random}. Each data point in a health record reflects a conscious decision to collect information, based on the patient's condition and the clinical staff's assessment of what was needed and when. This observation process of interactions, which we refer to as \emph{clinical presence} in the medical context, is informative as missingness and observation times are not independent of the patient's conditions. 

When considering clinical presence in the prediction modelling context, current modelling strategies leverage some measures of this process to improve predictive performance~\cite{sisk2020}. Indicators of missing tests, number of visits, and time since the last visit have been used as proxies for the patient's condition, as temporality and missingness patterns often reflect the severity of a patient's condition, but these proxies are ad hoc by nature and do not fully reflect the data-generating process. Instead of these approaches, practitioners can jointly model the clinical presence with the outcome process. Joint models~\cite{sperrin2017informative, gasparini2020mixed, su2019sensitivity}, marked point process~\cite{islam2017marked} and Markov models~\cite{alaa2017learning} model the visit process and the outcome of interest. Through shared effects, the data-generating process informs the modelling of the outcome of interest. However, these methods either rely on correct model specification with strong parametric assumptions or do not scale well to large datasets. Our work relies on neural networks to address these limitations.


Critically, clinical presence processes differ across hospitals~\cite{johnson2018generalizability}, regions, and countries, as medical training and practice may differ and may be subject to policies~\cite{subbaswamy2020development}, insurance incentives, and the evolution of medical knowledge~\cite{ghassemi2019practical}. 
This \textit{clinical presence shift} highlights a crucial limitation of clinical prediction models trained on EHRs; they may embed clinical presence patterns that do not transport under shift as practices and policies evolve~\cite{van2020cautionary}. Given the risk associated with the decisions these models inform, improving models' transportability under these naturally occurring shifts is critical to applications in healthcare. We are unaware of work that has explored these models' transportability under clinical presence shifts. Our work fills this gap by examining how handling irregular timings in longitudinal data may impact models' transportability under clinical presence shifts. When considering potential shifts, modelling pipelines \textit{discard} clinical presence patterns with the goal of improving transportability between observational settings. Although practitioners justify this practice to avoid leveraging a changeable and unpredictable process~\cite{van2020cautionary, xiao2018opportunities} in the hope of improved transportability, evidence to back up or substantiate this claim in practice is lacking. On the contrary, we show that predictive performances decrease under shift when clinical presence is ignored.

As a solution for improved transportability under clinical presence shifts, we propose a recurrent neural network that jointly models the time-to-event outcome of interest, inter-observation times and missingness features of a multivariate longitudinal process. Specifically, we model the time-to-event outcome using a neural network extension of the Cox model~\cite{cox1972regression}, known as DeepSurv~\cite{katzman2018deepsurv}, and, in parallel, the time to the subsequent observation and the missingness in laboratory tests of the multivariate longitudinal process. The proposed model, referred to as DeepJoint, enables scalable joint modelling by maximizing the full likelihood through end-to-end gradient descent.


Our work is motivated by the prediction of mortality for patients admitted to intensive care, using data from the Medical Information Mart for Intensive Care III (\textsc{Mimic III})~\cite{johnson2016mimic}. We use laboratory tests performed during the first 24 hours following admission as risk predictors for mortality after 24 hours. Of particular interest is the shift in the clinical presence patterns between patients admitted on weekends and weekdays. Through this natural experiment, our work explores how joint modelling improves the transportability of a model trained on patients admitted on weekdays to weekends. Our experiments show that the additional modelling of the time between successive orderings of laboratory tests and which out of a suite of 21 possible laboratory tests have been performed improves robustness to this shift in the observation process. Contrary to prevailing views, this experiment shows that not accounting for clinical presence not only reduces predictive performance but also harms transportability. By explicitly modelling clinical presence, DeepJoint results in a more transportable clinical prediction model in settings where the clinical presence process is likely to evolve. 

 The outline of the paper is as follows. In Section~\ref{sec:background}, we provide an introduction to the machine learning concepts used in the paper. In Section~\ref{sec:literature}, we review existing strategies to model clinical presence and improve transportability under shift. In Section~\ref{sec:cp}, we define the concept of clinical presence shift and consider transportability to different settings. In Section~\ref{sec:methodology}, we propose the DeepJoint architecture and describe the training (estimation) of the model. In Section~\ref{deepjoint:sec:experiment}, we present the results of the application to the \textsc{Mimic III} dataset, and we conclude with a discussion in Section~\ref{sec:discussion}.   

\section{Background}
\label{sec:background}

We first review the neural network components of our proposed architecture, which might be unfamiliar to statistical readers. After a brief introduction to neural networks, we describe the recurrent neural network structure used for handling longitudinal observations, the monotonic neural network, which we use to capture information about the time between successive observations (the inter-observation time), and the DeepSurv method, which we use for the survival outcome of interest. For an exhaustive description of neural networks and their evolution, please refer to~\cite{bishop2006pattern, goodfellow2016deep}. 

\subsection{Neurons}
Neural networks aim to mimic biological brains composed of simple processing units --- neurons --- that, together, form complex systems. Formally, given a multidimensional \emph{input} $x \in \mathbb{R}^m$ with $m \in \mathbb{N}$ its dimension, a neuron consists of a linear 
combination of its inputs --- defined by a \emph{weight vector} $W \in \mathbb{R}^m$ and a bias\footnote{While this denomination is common in the ML literature, this term corresponds to the statistical concept of \emph{intercept}. To avoid confusion, we refer to both weights and bias as a neuron's \emph{parameters} throughout.} $b \in \mathbb{R}$ that weigh the neuron's input connections --- which is then passed through a non-linear \emph{activation function} $\phi$ responsible for quantifying the neuron's response into an \emph{output} response. The neural output $o(x)$ can be expressed as:
$$o(x) = \phi(x^T W + b)$$

The activation function determines whether an input triggers the associated neuron. Mathematically, this component ensures neural networks' non-linearity, and, consequently, their modelling flexibility. In this work we use the SoftPlus~\cite{dugas2000incorporating} function defined as $\textrm{SoftPlus}(x) = \ln(1 +e^x)$. The motivation behind SoftPlus is to unconstrain the output values, and to increase the range of non-zero gradient~\cite{szandala2021review} --- critical for avoiding the problem of vanishing gradient in training. 

There exist activation functions that consider the outputs of multiple neurons simultaneously. Critically, these functions quantify a neuron's response relative to other neurons' activation. For instance, the SoftMax function~\cite{goodfellow2016deep} ensures the positivity and the summation to $1$ of its outputs, through the transformation:
$$\text{SoftMax}(z_i) = \frac{e^{z_i}}{\sum_{j=1}^n e^{z_j}}$$
with $z_i$ the output of neuron $i$ out of the $n$ considered. This function outputs the activation probability distribution over the set of input neurons.

Individually, neurons resemble generalised linear regressions with the activation corresponding to the inverse of the link function. Interconnected, neurons form a network with complex behaviours. The next sections describe two types of neural architectures tackling static and temporal data: Multi-Layer Perceptrons and Recurrent Neural Networks~(RNN).

\subsection{Multi-layer perceptrons}
\begin{figure*}[!ht]
	\centering
	\includegraphics[width=0.8\textwidth]{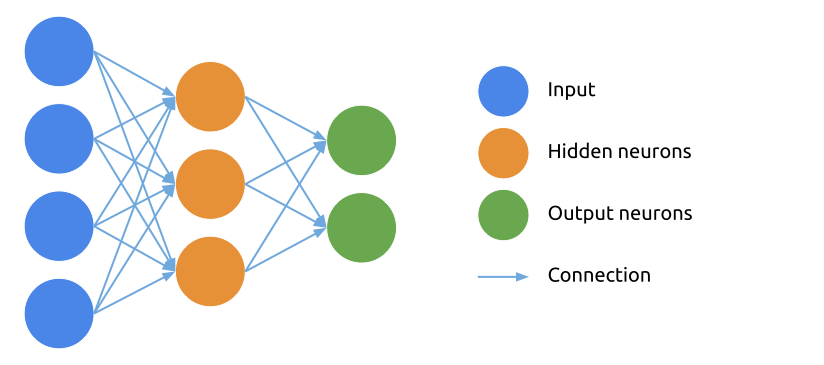}
	\caption{Fully connected neural network with one hidden layer.}
	\label{background:fig:vanilla}
\end{figure*}

A simple arrangement of independent neurons with the same inputs is known as a \emph{layer}. Accumulating multiple layers results in a multi-layer perceptron. Inner layers $l$ are referred to as \emph{hidden} if their outputs serve as inputs to another layer of neurons. 
For prediction, the input data is processed by each neuron of the first layer, forming outputs that are then used by the following layer as inputs. This process is iterated until reaching the final \emph{output layer} that returns the predicted outcomes. This procedure from the inputs to the output layer is known as the \emph{forward propagation}. For instance, consider a two-layer network as presented in Figure~\ref{background:fig:vanilla}: a hidden layer takes the data as input, transforms it, and passes it to the output layer. This last layer aims to generate the target labels. A central property of this simple arrangement is its ability to approximate \emph{any} continuous functions \cite{hornik1989multilayer}.

\subsection{Recurrent neural network}
\begin{figure*}[!ht]
	\centering
	\includegraphics[width=0.9\textwidth]{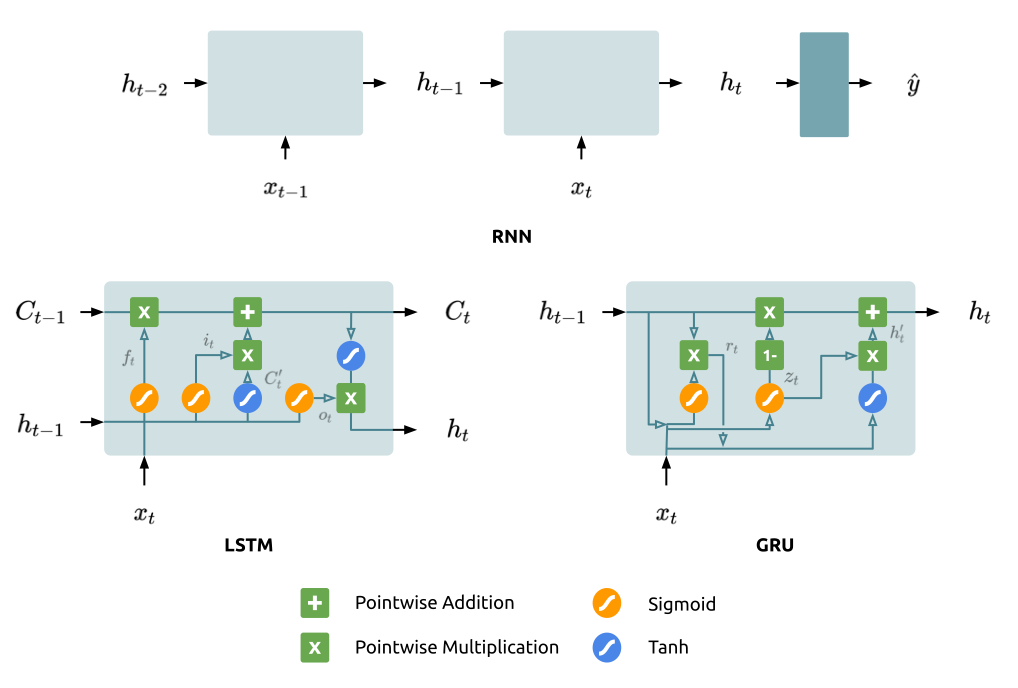}
	\caption{Architectures of a Recurrent Neural Network with LSTM architecture.}
	\label{background:fig:rnns}
\end{figure*}

When considering repeated measurements, RNNs take advantage of the sequential nature of the data. At each new time point, the model is applied to the new observation, while also considering past information through a \emph{memory} state. Formally, a RNN, $\Psi$, is applied on the observation $x_t$ observed at time $t$ using the memory state $h_{t - 1} \in \mathbb{R}^d$ obtained at time $t - 1$ with $d$ the embedding dimension, corresponding to the number of neurons used for representing the time series. This defines the recurrence: $$h_t = \Psi(x_t, h_{t-1})$$
An output layer can then be employed to predict the outcome $\hat{y}$ from any of the memory states, as shown in Figure~\ref{background:fig:rnns}.

While simple, this iterative model presents a limited capacity to capture long-term trends~\cite{hochreiter1997long}. This phenomenon stems from the training procedure that updates the model's parameters given the gradient of the loss. The gradient associated with earlier events tends to fade at each new time point. This vanishing gradient favours recent events in the parameters' updates. 
Alternative architectures tackle this particular issue by adding connections to maintain the gradient's strength over time. Two such architectures used in this work are the Long Short Term Memory ({LSTM}~\cite{hochreiter1997long}) and the Gated Recursive Unit ({GRU}~\cite{cho2014properties}) architectures. These recursive models, presented in Figure~\ref{background:fig:rnns}, differ from the simple {RNN} in their updating mechanisms of the memory state via multiple combinations of neurons and activation functions that balance the influence of the previous memory state and the new observation at each new time point\footnote{We invite the reader to refer to the original works~\cite{cho2014properties, hochreiter1997long} for detailed descriptions of these mechanisms.}. These learnt mechanisms choose which information should be kept from a new observation and from the long-term memory. Specifically, the {LSTM} architecture introduces a new memory state $C$, which represents this long-term memory. GRU presents a similar idea in a simplified form by using a single memory state balancing both long-term and short-term information.

Mathematically, the LSTM can be described as the following set of equations:
\begin{align}
	f_t &= \sigma(W_f x_t + U_f h_{t-1}) \tag{Forget Gate}\\
	i_t &= \sigma(W_i x_t + U_i h_{t-1}) \tag{Input Gate}\\
	y_t &= \sigma(W_y x_t + U_y h_{t-1}) \tag{Output Gate}\\
	C_t' &= \text{TanH}(W_c x_t + U_c h_{t-1})\notag\\
	C_t &= f_t \cdot C_{t-1} + i_t \cdot C_t' \tag{Memory State}\\
	h_t &= y_t \cdot \text{TanH}(C_t)\notag
\end{align}
where $W_g$ and $U_g$ denote the set of parameters associated to the gate $g$, and $\sigma$ is the logistic function, TanH is the hyperbolic tangent function -- two common \textit{activation functions} used to introduced non-linearity in neural networks. Similarly, the GRU is defined as:
\begin{align}
    z_t &= \sigma(W_z x_t + U_z h_{t-1}) \tag{Update Gate}\\
    r_t &= \sigma(W_r x_t + U_r h_{t-1}) \tag{Reset Gate}\\
    h_t' &= \text{TanH}(W x_t + r_t \cdot U h_{t-1}) \tag{Update}\\
    h_t &= (1 - z_t) \cdot h_{t-1} + z_t \cdot h_t' \notag
\end{align}

\subsection{Monotonic neural network}
Neural networks can be organised to enforce constraints to ensure a desired property. One such subtype of \emph{constrained} neural networks is the monotonic neural network. A monotonic neural network is a neural network that ensures the monotonicity of its output, given its inputs. Formally, a monotonic network $\Psi$ presents an increased output for any increase in the input: 
$$\forall (t, t') \in \mathcal{X}^2, t' \geq_\mathcal{X} t \implies \Psi(t') \geq_\mathcal{Y} \Psi(t)$$
given the ordering relations $\geq_\mathcal{X}$ and $\geq_\mathcal{Y}$ for the input space $\mathcal{X}$ and output space $\mathcal{Y}$.
An efficient way to achieve this is to constrain all neurons' weights to be positive by applying a transformation over them, such as the absolute value~\cite{omi2019fully} or square function~\cite{ rindt2022survival, chilinski2020neural, jeanselme2023neural}. This last transformation redefines a neuron's output as $y(x) = \phi(x^T W^\mathbf{2} + b)$, where $W^\mathbf{2}$ denotes the vector obtained by squaring each entry of $W$ element-wise.. The square function's differentiability improves the training convergence compared to the non-differentiable absolute value. Crucially, \cite{daniels2010monotone, lang2005monotonic} demonstrate that positively weighted networks are universal monotone approximators, which ensures the model's flexibility to model any monotonic function and motivates our use of monotonic neural networks for modelling the cumulative incidence function in survival analysis.

\subsection{DeepSurv}

Neural networks have been proposed to extend traditional survival models~\cite{katzman2018deepsurv, lee2018deephit, jeanselme2023neural}. Of particular interest in our work, 
DeepSurv \cite{katzman2018deepsurv} adapts the Cox model with non-linear covariate interactions. Traditionally, the Cox proportional hazards model~\cite{cox1972regression} is composed of a linear combination of covariates parameterised by a vector $\beta$: $\eta(x) = \beta^T x$, to model deviations from a population's non-parametric baseline hazard function $\lambda_0(t)$, which can be recovered via a Breslow estimator. Formally, this approach models the hazard as: $$\lambda(t\mid X) = \lambda_0(t) e^{\eta(X)}$$ 

The resulting proportionality assumption allows the parameters associated with $\eta$ to be estimated independently of the non-parametric baseline hazard function, by maximising the partial log-likelihood defined as:
$$l_{\text{CoxPH}} = \sum_{i, d_i = 1} \left[ \eta(x_i) - \log \sum_{j, t_j \geq t_i} e^{\eta(x_j)} \right]$$

DeepSurv~\cite{katzman2018deepsurv} proposes to make $\eta$, an output of a multi-layer perceptron, $\Psi$, i.e. $\eta(x) = \Psi(x)$. Training then follows the same procedure as the traditional Cox model, relying on maximisation of the partial likelihood and a Breslow cumulative hazard estimator.

\subsection{Neural network fitting}
Training a neural network consists of estimating the optimal values of each parameter --- weights and bias associated with each neuron --- to minimise the difference between the predicted outputs $\hat{y}$ and target values $y$. Formally, one aims to optimise a loss function $\mathcal{L}$ which quantifies the error between the predicted value and the observed value. The dependencies between neurons make this optimisation particularly complex as one can not optimise for each parameter independently. Training a model requires finding the set of parameters that globally, not locally, solves the optimisation problem. An additional challenge comes from the model's non-linearity, which leads to the non-convexity of the loss given the parameters. 


To tackle these challenges, our work relies on gradient descent optimisation. Gradient descent minimises the loss function by iteratively updating the parameters' values in \emph{the opposite direction of the gradient of the loss}. While gradient descent is guaranteed to recover the global minimum with a convex loss, it may converge towards local minima when this property does not hold. To avoid converging towards local minima, stochastic approaches encourage exploration beyond the current parameters' selection. The strategy adopted in our work is the Stochastic Gradient Descent. This algorithm estimates the gradient on a subset of the data, randomly drawn at each iteration of the algorithm, instead of using all training data. The sample size used to approximate the gradient is selected as a hyperparameter known as batch size. This estimation reduces the training computational cost, improving convergence speed.

\section{Related work}
\label{sec:literature}
In the literature, multiple approaches have been introduced to model clinical time series and to tackle potential distribution shifts. In the following, we review these works.

\subsection{Irregular timing modelling}
In healthcare, observations rarely occur following a precise schedule, and practitioners only order a subset of the medical tests at each encounter, particularly in the intensive care settings. How to handle these temporal and missingness dimensions is an active research question in the statistical literature (see~\cite{sisk2020informative} for a detailed review). In this work, we divide the methodologies used in the literature into three categories: (i)~Preprocessing, (ii)~Featurised, and (iii)~Jointly modelled.

\paragraph{Preprocessing.} Practitioners deal with irregular timings as a nuisance at preprocessing time. First, if practitioners assume that the observed covariate distribution $q(X^*)$ is a representative sample of the underlying observable covariate distribution $q(X)$, with $X$ the observable covariates and $X^*$, the observed ones, one can ignore these timings. For example, regular re-sampling and Missing Completely At Random / Missing at Random imputation reflect this assumption. If incorrect, this can lead to biased estimates and potentially misleading conclusions. Second, if practitioners recognise these timings are informative, they aim to recover $q(X)$ from the observed $q(X^*)$ to then leverage the true underlying covariate trajectory. \cite{lipton2016directly} underline how difficult it is to adequately reverse this observation process and question the utility in recovering the underlying distribution of this second family of methods.

\paragraph{Featurised.} In this setting, practitioners assume that irregular timings are not nuisances to address but proxies to the patient's condition. One can extract measures of clinical presence and use them as inputs to their models of interest. For instance, the use of missing indicators~\cite{lipton2016directly}, observation times~\cite{agniel2018biases, che2018recurrent, sousa2020improving}, inter-observation time~\cite{cai2018medical, choi2016doctor, choi2016retain, moskovitch2015outcomes, zhang2019attain}, or frequency~\cite{pivovarov2014identifying} has improved clinical prediction models' performance. 

\paragraph{Jointly modelled.} Instead of the approaches described above, practitioners can directly model these irregular timings. Joint models~\cite{gasparini2020mixed, sperrin2017informative, su2019sensitivity}, marked point process~\cite{islam2017marked} and Markov models~\cite{alaa2017learning} jointly model the visit process and the outcome of interest. Typically, through shared random effects in joint models, the observation process informs the modelling of the outcome of interest. However, these methods rely on strong parametric assumptions and generally do not scale to large datasets. Analogous ML approaches to those statistical approaches have been proposed for modelling missingness~\cite{twala2008good} or irregular timings~\cite{weerakody2021review}. These ML methods aim to integrate the temporal information by slowly converging to a \textit{stable} latent state as time passes through a decay parameter~\cite{baytas2017patient, che2018recurrent, pham2016deepcare} or multi-level memory~\cite{mozer2017discrete}. For instance, {GRU}-D~\cite{che2018recurrent} has shown promising results by extending the Gated Recurrent Unit ({GRU}) with an exponential decay on the hidden state, incorporating the temporal dimension in the embedding. 

While these approaches tackle the problem of irregular timings, they rely on different assumptions regarding clinical presence informativeness, usually focusing on one dimension of the process. To our knowledge, no study has evaluated their sensitivity to distribution shifts.

\subsection{Distribution shifts}
\label{dj:sec:shift}
Predictive performances often degrade at deployment. This phenomenon is explained by a drift in the joint distribution $q(X, Y)$ of covariates $X$ and outcomes $Y$ between training and deployment. This phenomenon, known as distribution shift~\cite{quinonero2008dataset}, has been extensively studied in the literature~\cite{moreno2012unifying, zhang2013domain} with two subtypes receiving particular attention:
\begin{enumerate}
    \item \textit{Covariate shift.} The covariate distribution varies $q_{train}(X) \neq q_{test}(X)$ while the conditional relation $Y$ given $X$ remains  the same between training and deployment. Depending on the assumed generative process ($X \leftarrow Y$ or $X \rightarrow Y$), the issue of target shift, $q_{train}(Y) \neq q_{test}(Y)$, has also been similarly investigated in the literature.
    \item \textit{Concept shift.} The covariate distribution remains stable but the conditional distribution evolves $q_{train}(Y \mid X) \neq q_{test}(Y \mid X)$.
\end{enumerate}

Extensive literature exists on detection and mitigation strategies to tackle these shifts. These aspects are beyond the scope of this paper. Readers are referred to existing reviews~\cite{lu2018learning, nair2019covariate} for further information. We focus, in what follows, on the contributions that shaped our proposed approach and study.

\paragraph{Detecting shift.} A first step towards addressing distribution shifts is to be alerted to a model's unreliability. The similarity between the covariate distributions in the training and deployment datasets~\cite{park2021reliable} may provide evidence of a distribution shift. However, when only a sample is available from the deployment distribution, practitioners must rely on either anomaly detection~\cite{chandola2009anomaly} to identify how a given point differs from the original distribution or favour models with outputs that capture this uncertainty~\cite{hendrycks2016baseline, lee2018simple}.

\paragraph{Domain adaptation.} When samples from the deployment distribution are available, one may alter the model's training to better model the deployment distribution. For instance, \cite{zhang2013domain} describe an inverse weighting strategy to improve modelling under the two previously described shifts. Similarly, \cite{fang2020rethinking} further adapt inverse weighting to improve the transportability of deep learning models. \cite{lipton2018detecting} propose a strategy when one does not have access to labels in the target domain. However, all these strategies assume access to samples from the deployment distribution.

\paragraph{Domain generalisation.} In contrast to domain adaptation, domain generalisation aims to create models transportable under shifts \textit{without access to samples from the deployment distribution}. \cite{zhou2022domain} review the literature on this topic. Closest to our work, regularisation and self-supervised learning techniques~\cite{mohseni2020self} may improve the transportability of a model to a new domain. For instance, \cite{mohseni2020self} propose image reconstruction as self-pretraining, resulting in detecting out-of-distribution images. Our work resembles this self-supervision as we aim to train the architecture to model the observation process.

\paragraph{Domain shifts in medicine.} Despite the extensive literature on this topic, existing methods assume specific shifts or access to samples from the deployment distribution, not reflecting the nature of real-world medical shifts~\cite {spathis2022looking}. \cite{spathis2022looking} criticise this lack of real-world shift studies and demonstrate that multiple state-of-the-art strategies do not improve out-of-distribution prediction across hospitals of the e{ICU} dataset~\cite{pollard2018eicu} on in-hospital mortality prediction. Similarly, \cite{guo2022evaluation} show that domain generalisation or adaption techniques fail to improve transportability in the \textsc{Mimic} dataset across years. In this same setting, \cite{nestor2019feature} propose extracting clinically-relevant features to improve transportability. The authors demonstrate these features reduce the necessary modelling complexity, with logistic regression presenting one of the best performances and transportability. Note that while these works study mortality prediction, only \cite{pfisterer2022evaluating} explore the transportability of survival models and show limited gain from distribution shift mitigation strategies.

A final challenge in clinical distribution shifts is the lack of publicly available benchmarks with limited data across hospitals --- with eICU being a rare exception with multiple institutions --- and limited common measurements to evaluate shifts. In this work, we introduce a novel way to assess transportability through natural experiments, such as the weekend effect, present in medical datasets.

\section{Clinical presence shift}
\label{sec:cp}
While prior works have examined different types of distribution shifts and proposed mitigation strategies, we focus on a distinct and underexplored form of distribution shift: clinical presence shift. This shift arises from changes in clinical practice that alter when and what patient information is recorded. This distribution shift is the core motivation of our work, as such shift can threaten the generalizability of clinical prediction models trained under specific clinical presence patterns.

\begin{figure}[ht]
	\centering
	\includegraphics[width=\textwidth]{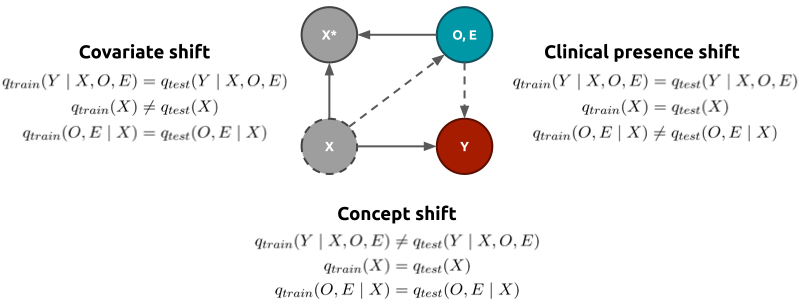}
	\caption{Clinical presence shift. Full-circled covariates are observed, and dashed ones are potentially unobserved. Dashed arrows indicate potential dependencies, while the solid ones show assumed dependencies. Distinct from covariate and concept shift, clinical presence shift is a shift in the joint distribution $O, E$ while all other distributions remain the same.}
	\label{dj:fig:shift}
\end{figure}

Clinical presence shift can be understood as a combination of both covariate and concept shifts with regard to the observed distribution. Figure~\ref{dj:fig:shift} redefines the traditional problem of distribution shifts by adding the observation process characterised by $(O, E)$, with $E$, the time since the last encounter and $O$, the indicator vector of covariates observed: one dimension per covariate. In this figure, we dissociate observed $X^*$ and observable $X$ covariates and introduce their dependencies. Additionally, we display potential dependencies between unobserved covariates and the observation process that would reflect \textit{not at random} patterns, as well as one between the observation process and observed outcomes, indicating the possible impact of the observation on the outcome itself. This representation highlights a new type of shift: a \emph{clinical presence shift}. Assuming an unchanging concept distribution, $q(Y\mid X, E, O)$ and covariate distribution, $q(X)$, a clinical presence shift is a change in the distribution $q(E, O)$ resulting in shifts in the observed covariate distribution $q(X^*)$ and the modelled $q(Y\mid X^*)$. Generally, the problem of clinical presence shift can be seen as both a concept and covariate shift as both joint distributions change. In the clinical setting, this means that even when faced with the same underlying population with the same associated outcome, a model trained under a given observation process may not be adapted to a new one. 

\section{Joint modelling for prediction under clinical presence}
\label{sec:methodology}

The proposed method is at the intersection of two fields of research: joint modelling and multitask learning. Firstly, joint models have been proposed in the statistical literature to incorporate informative processes into the outcome model through shared random effects~\cite{gasparini2020mixed, sperrin2017informative, su2019sensitivity}. These models suffer from poor scalability with increasing sample size and number of longitudinal outcomes. Our proposed deep learning architecture tackles this issue and relaxes the parametric assumptions often made by these models. Secondly, our method is also inspired by the multitask learning literature in which models aim to predict multiple outcomes to improve performance~\cite{liu2020multi, caruana1997multitask, caruana1998multitask}. Our work bridges the gap between these two domains by using multitask learning to model clinical presence, forcing the shared embedding to contain a representation informative of both the observation process and outcome of interest.

\subsection{DeepJoint architecture}
\begin{figure}[ht]
	\centering
	\includegraphics[width=\linewidth]{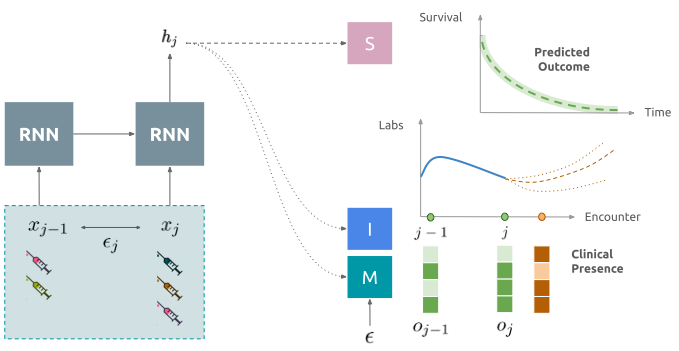}
	\caption{DeepJoint Model - Multitask modelling of clinical presence and survival outcome. A {RNN} extracts an embedding $h_j$ then used to model clinical presence through the networks \textbf{M} --- to model the missingness patterns, \textbf{I} --- to model the inter-observation times, and the survival outcome through the network \textbf{S}}.
	\label{dj:fig:model}
\end{figure} 

The proposed architecture, illustrated in Figure~\ref{dj:fig:model}, models two dimensions of clinical presence: inter-observation times and missingness, modelled through two different neural networks. Each relies on the embedding $h_{j}$ outputted by a RNN with input $(x^*_j, o_j, \epsilon_j)$ at encounter $j$ where $x^*$ are observed covariates -- in our work, laboratory tests, $o$ is the indicator of observation and $\epsilon$, the inter-observation time. Note that, by training the {RNN}, we embed in $h_j$ all observed observations up to and including $j$, corresponding to $\mathcal{H}_{j + 1}$. In our implementation, we adopt the commonly used {LSTM}~\cite{hochreiter1997long} to model the sequential nature of the observed data and capture temporal dependencies. Note that the {LSTM}'s inputs are the series of laboratory tests at \emph{irregular times}. The resulting shared embedding at the end of the 24-hour observation period is inputted to the survival model. The following describes the implementation choice for each of these components.



\paragraph{Temporal process: When will the subsequent measurement take place?} A monotonic positive neural network $\mathbf{I}$ models the cumulative hazard associated to the time to the next measurement. As previously described, we enforce (i)~monotonicity by using the square of all weights and (ii)~postivity through a final Softplus layer. Formally, the network models:
$$\mathbb{P}(E_{j + 1} < \epsilon \mid \mathcal{H}_{j + 1}) = 1 - \exp(- \mathbf{I}(h_{j}, \epsilon))$$

\paragraph{Missingness process: Which laboratory tests will be performed next?} A multi-layer perceptron $\mathbf{M}$ uses $h_{j}$ as inputs to model the probability of observing the different covariates at encounter $j + 1$, i.e.~which tests are likely to be performed by $\epsilon$. 
$$\mathbb{P}(O_{j + 1} = o \mid E_{j + 1} < \epsilon, \mathcal{H}_{j + 1}) = \mathbf{M}(h_{j}, \epsilon)$$

\paragraph{Survival outcome: How long will the patient survive?} Finally, to model the hazard function $\lambda(t \mid \mathcal{H}_{l_i + 1})$ at the last encounter $l_i$ in the observation period, we use the DeepSurv model. Given $\lambda_0(.)$, the baseline hazard, the hazard of observing an event at horizon $t$ is expressed as:
$$\lambda(t \mid \mathcal{H}_{l_i + 1}) = \lambda_0(t) \exp(\mathbf{S}(h_{l_i})) $$
The final model, therefore, combines existing architectures in a novel way to model clinical presence and survival outcomes jointly. This results in a latent representation $h_j$, which embeds both the observation process and survival outcome.

\subsection{Training}
Training this model consists of maximising the sampling process and survival likelihoods. The average loss of the survival, temporal and missingness processes is backpropagated. Following the multitask literature, we ensure that the inter-observation time and missingness are balanced using a dynamic weighting average scheme~\cite{liu2019end}, as imbalances can significantly impact performance~\cite{gong2019comparison}. Specifically, we weight these two modelling tasks using the relative change at iteration $s$ as follows:
$$\forall\;task \in \{I, M\}, w_{task}(s) = \frac{L_{task}(s)}{L_{task}(s - 1)\cdot \theta},$$ which is then normalised between the two clinical presence dimensions using a Softmax. $L_{task}$ is the average training likelihood for the given task, and $\theta$ is a temperature hyperparameter that controls softness, i.e.~larger values would lead to equal weights. In this work, we follow the empirical recommendation of the original paper and set $\theta$ to $2$.


\paragraph{Temporal process.} Our choice of monotonic networks results in the exact computation of the likelihood associated with the temporal process. The model outputs the cumulative intensity at horizon $\epsilon$ and automatic differentiation results in the instantaneous intensity.
$$l_I = \sum_i \frac{1}{l_i - 1} \sum_{j \in [\![1, l_i-1]\!]} \left( \mathbf{I}(h_{i, j}, \epsilon_{i, j+1}) - \log \frac{\partial \mathbf{I}(h_{i, j}, \epsilon)}{\partial \epsilon}\bigg|_{\epsilon = \epsilon_{i, j+1}} \right)$$

\paragraph{Missingness process.} Similarly, the likelihood for the associated mark for the missingness process results in the binary cross entropy loss:
$$l_M = - \sum_i \frac{1}{l_i - 1} \sum_{j \in [\![1, l_i-1]\!]} \big( (1 - o_{i, j+1})\cdot \log[\mathbf{M}(h_{i, j}, \epsilon_{i, j+1})] + o_{i, j+1}\cdot \log[1 - \mathbf{M}(h_{i, j}, \epsilon_{i, j+1})] \big)$$

\paragraph{Survival outcome.} As previously described, DeepSurv~\cite{katzman2018deepsurv} uses a neural network to express the log-hazard in a Cox model. This means that the hazard shift is fully characterized by the embedding $h$, which represents all the previous laboratory tests. In this context, the neural network's parameters are optimised using the partial log-likelihood:
$$l_S = \frac{1}{\sum_{i, d_i = 1} 1} \sum_{i, d_i = 1} \left( \mathbf{S}(h_{i, l_i}) - \log \sum_{k, t_k > t_i} \exp(\mathbf{S}(h_{i, l_i})) \right)$$
Finally, the population baseline hazard, $\lambda_0(t)$, is estimated with a Breslow estimator.

The final loss $l$ is defined at iteration $s$ as $$l(s) = (1 - \alpha)\cdot l_S + \alpha \sum_{task \in\{I, M\}} w_{task}(s)\cdot l_{task}$$
with $\alpha \in [0, 1]$, a hyperparameter balancing the survival and observation process losses. This loss is computed on the training set and backpropagated throughout the architecture. We also compute it on a validation set for early stopping of the multitask training with $w_{task} = 1$. Then, we fine-tune each task-specific network, not backpropagating through the shared {RNN}. This corresponds to training each task-specific network further by maximising the task-specific likelihood without updating the shared representation.

\subsection{Theoretical transportability}
\label{deepjoint:proof}

In this section, we theoretically justify why the proposed architecture is robust to clinical presence shift. 
\cite{mao2020multitask} show that, under small shifts in inputs, multitask learning improves transportability. Specifically, the authors demonstrates the proportionality of the expected error to the inverse of the number of outcomes modelled under small perturbations of the input. 

\paragraph{Intuition.} The following theorem quantifies the error resulting from a distribution shift between two time series. When the difference between these time series is negligible, the theorem states that the more outcomes a multitask neural network models, the smaller the difference between the predictions associated with the shifted and true time series. This means that the model would be more robust to small shifts in the input time series. Note that this result relies on three assumptions: (i)~the gradients are i.i.d., which requires the time series at deployment to be i.i.d., (ii)~the average gradient is null, corresponding to the training having converged, and (iii)~the perturbation is infinitesimal. The distance between time series is challenging to measure and unlikely to be negligible between patients in different medical settings; this means that this last assumption is less likely to be met. However, this theorem provides an intuition into why modelling clinical presence as additional outcome tasks of the proposed DeepJoint may improve transportability under small shifts in clinical presence.

\paragraph{Formalisation.} First, we introduce the \textit{clinical presence vulnerability} defined as the expected change in loss under two different observation processes. Consider the observed time series at development $\mathcal{H}$, and at deployment $\mathcal{H}'$, the target outcomes $y$, a loss $\mathcal{L}$, such that the difference between the marked point processes resulting from the two observation processes lies in a p-norm ($||\cdot||_p$) bounded ball with radius $r$, i.e.~$||\mathcal{H} - \mathcal{H}'||_p < r$. Note that we assume the existence of a p-normed time series space, which includes an operator $+$, where $+\ \delta$ describes a perturbation of the time series. Under this assumption, the clinical presence vulnerability over the target dataset is:
$$\Delta \mathbb{E} := \mathbb{E}\big[|\mathcal{L}(\mathcal{H}, y) - \mathcal{L}(\mathcal{H}', y)|\big] \leq \mathbb{E}\big[\max_{||\delta||_p < r} |\mathcal{L}(\mathcal{H}, y) - \mathcal{L}(\mathcal{H} + \delta, y)|\big]$$
Assuming \emph{infinitesimal perturbations} in the observation processes, i.e.~$r \rightarrow 0$, one can develop the upper-bound using Taylor expansion: 
$$|\mathcal{L}(\mathcal{H}, y) - \mathcal{L}(\mathcal{H} + \delta, y)\mid = |\partial_{H} \mathcal{L}(\mathcal{H}, y)\delta + \mathcal{O}(\delta)|$$
with $\partial_{H}\mathcal{L}$ the partial derivative of the loss given the input time series $H$.
Using the property of the dual norm $d$, one can show that:
$$\Delta \mathbb{E} \propto \partial_H \mathbb{E}\big[||\mathcal{L}(\mathcal{H}, y)||_d\big]$$

\begin{theorem}[Multitask robustness from~\cite{mao2020multitask} - see proof in \cite{mao2020multitask}]
	\label{theorem}
	If the modelled outcomes are correlated with each other such that the covariance between the vector of gradients of the loss for outcome $u$ and outcome $v$ is $Cov(g_u, g_v)$, and these gradients are i.i.d. with zero for the average, then:
	$$\Delta\mathbb{E} \propto \sqrt{\frac{1 + \frac{2}{M} \sum_{u = 1}^{M} \sum_{v = 1}^{u - 1} \frac{Cov(g_u, g_v)}{Cov(g_u, g_u)}}{M}}$$
	where M is the number of the modelled outcomes and $g_u$, the vector of gradient defined as $\forall \mathcal{H}, \partial_H \mathcal{L}_u(\mathcal{H}, y)$ associated with the task-specific loss associated with the time series $\mathcal{H}$.
\end{theorem}

\section{Case study: Predictive modelling under weekend effect shift}
\label{deepjoint:sec:experiment}
Using laboratory tests performed during the 24-hour observation period after admission to the {ICU} in the Medical Information Mart for Intensive Care III (\textsc{Mimic III})~\cite{johnson2016mimic} dataset, we modelled survival post 24-hour of ICU admission. As in the publicly available version of the dataset, dates have been shuffled for anonymisation and data gathered in a single hospital, it was not possible to study transportability across time and place. However, as admission days have been maintained, we investigated how model performance would change under the natural experiment of weekend effect~\cite{aylin2015making, pauls2017weekend}, where a distribution shift occurs between patients and clinical care over weekends and weekdays. Specifically, we evaluated how a model trained using patients admitted on weekdays would perform in the population admitted during weekends. For reproducibility, the proposed models' -- including DeepJoint -- and experiments' implementations are available on Github\footnote{\url{https://github.com/Jeanselme/MultitaskTransportability}}.

\subsection{Dataset}
The \textsc{Mimic III} dataset gathers laboratory tests, vital signs and diagnoses of 38,597 anonymised patients admitted to the Beth Israel Deaconess Medical Centre between 2001 and 2012. Our analysis focused on laboratory tests as they are more likely to reflect clinical expertise and, consequently, inform patient outcomes. This may not be true for other informative modalities which might present different clinical presence patterns,~e.g.~semi-automatic vital signs collection. After following the pre-processing from~\cite{wang2020mimic}, we selected a set of adults with shared laboratory tests using an ECLAT algorithm~\cite{zaki1997parallel} such that each laboratory test was performed at least once during the 24-hour post-admission, and patients survived this period. The resultant subset consisted of 31,692 patients with 21 different laboratory tests (See Tables~\ref{tab:mimic:pop} and~\ref{tab:mimic:labs} in Appendix for description).

\subsection{Empirical setting}

\paragraph{Baselines.} We compared \textbf{DeepJoint} against six alternative strategies for modelling clinical presence. All methods relied on the same normalised data imputed using the last observations carried forward with patient-mean imputation, i.e.~most recent laboratory tests replaced missing data; remaining missing tests used the lab overall mean.  Each approach extracts a representation from the longitudinal laboratory tests, then used by DeepSurv, a multi-layer perceptron, to estimate the hazard shift from the population hazard baseline.

The first two alternative strategies were non-longitudinal approaches:
\begin{itemize}
	\item \textbf{Last encounter (Last)}: Extract the last laboratory tests $l_i$ as representation for each patient $i$. This approach assumes no informativeness in the evolution of laboratory results, nor their temporality within the first 24 hours of ICU admission.
	\item \textbf{Summarising (Count)}: In addition to the last observed laboratory results, add the count of each test performed in the first 24 hours. This common practice utilises the counting process to reflect the severity of the patient's condition. Still, it ignores whether the patient is improving or worsening, as the temporal evolution is left aside. 
\end{itemize}

The remaining four strategies were {RNN}-based approaches that take advantage of the longitudinal evolution of laboratory tests. All these strategies use the laboratory test values as input to extract an embedding used by DeepSurv, as follows:
\begin{itemize}
	\item \textbf{Ignoring clinical presence (Ignore)}: An {LSTM} is trained on the imputed time series of laboratory tests, thus modelling the temporal order of input data but ignoring their irregularity and missingness patterns.
	\item \textbf{Resampling (Resample)}: Imputed laboratory tests are resampled every hour to satisfy the {LSTM} regularity assumption. At each hour of a patient's stay, data are averaged to obtain a time series of regularly sampled data. This approach ignores any informativeness of the observation process.
	\item \textbf{Modelling ({GRU}-D)}: The imputed laboratory tests are concatenated with missingness indicators and serve as inputs to a {GRU}-D model~\cite{che2018recurrent}. In comparison to the previous strategies, this RNN updates the embeddings as a function of the inter-observation times following a decaying function. The intuition is that the embedding, reflective of the patient health, stabilises with time. This decaying mechanism accounts for the temporality of the laboratory tests.
	\item \textbf{Featurization (Feature)}: Missingness indicators and time elapsed since previous observation~\cite{lipton2016directly} are two informative proxies of clinical presence. An {LSTM} uses for inputs both the laboratory tests and these proxies to model survival. This differs from DeepJoint as this architecture focuses on maximising the survival likelihood only, ignoring modelling of the clinical presence process.
\end{itemize}

The proposed \textbf{DeepJoint} architecture relies on the same input as Feature to meet the assumption of MAR. However, the proposed architecture differs in how the latent embedding at encounter $j$ is used to model the time to and the missingness at the subsequent encounter. Note that our approach is not dependent on a specific choice of {RNN}; any alternative to the proposed {LSTM} --- used for a fair comparison with Feature --- could be used for modelling longitudinal data.

\paragraph{Training procedure.} All methods used the same 80\%-20\% train-test patients split to train the different models. Their training relied on gradient backpropagation with an Adam optimiser~\cite{kingma2014adam} over 1000 epochs with early stopping on 10\% of the training set. We performed hyperparameter tuning using a 10\% left-aside set of patients from the training set on 50 random draws from the grid presented in Appendix Table~\ref{tab:grid}. For DeepJoint, the entire network was optimised for 500 epochs. The remaining iterations were for fine-tuning the survival network.

\begin{figure}[!ht]
	\centering
	\includegraphics[width=0.6\textwidth]{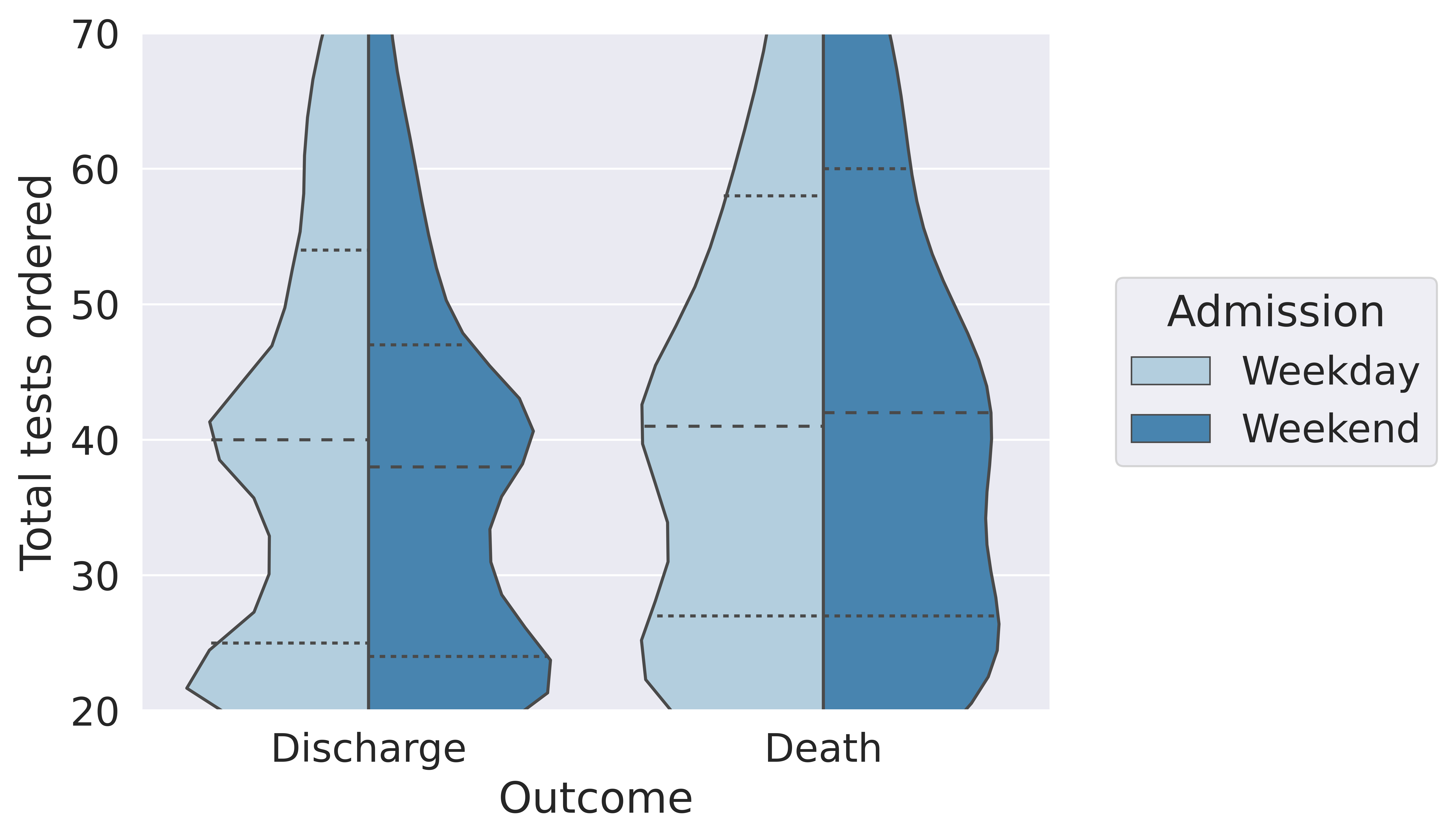}
	\caption{Total number of distinct tests ordered in the 24 hours after admission to the {ICU}. Patients admitted on weekends with favourable outcomes are less often tested than similar patients admitted on weekdays.}
	\label{fig:diff_weekday}
\end{figure}

The experiment described above explored how handling irregular timings impacts predictive performance. In a second set of experiments, we focused on how the day of admission impacts the observation process. Motivated by the difference in the counts of tests performed on patients admitted on weekends and weekdays, shown in Figure~\ref{fig:diff_weekday}, we hypothesised that clinical practice might differ between weekends and weekdays due to physicians and laboratory availability.

\begin{figure}[!ht]
	\centering
	\includegraphics[width=0.7\textwidth]{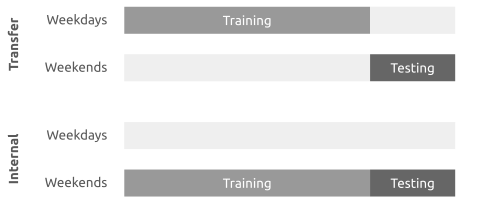}
	\caption{Transportability evaluation between patients admitted on weekdays and weekends. To ensure a fair comparison, all models are evaluated on the same population subset.}
	\label{fig:robust}
\end{figure}

The weekend effect is a well-known phenomenon in public health: patients admitted on weekends have worse outcomes than patients admitted on weekdays~\cite{pauls2017weekend}. While providing an opportunity to study survival models' transportability, this outcome gap represents a target shift under which we cannot directly compare performance. To make this comparison possible, we performed the following evaluation. We split patients given days of admission --- we define weekends admissions from Friday 8 pm to Sunday 8 pm. Each group was further divided between training and testing. As described in Figure~\ref{fig:robust}, a first model uses the train set of patients admitted on weekdays and tested on the test set of weekends-admitted patients. Then, a second model uses the training set of patients admitted during weekends and also tested on the test set of weekends-admitted patients. The two performance estimates are comparable as they were obtained on the same test set. The two resulting models are comparable as evaluated on the same test set. However, due to the difference in population size between patients admitted on weekdays compared to weekends, the quantity of data could confound the estimated transportability. To address this challenge, we followed the same subsampling approach as in~\cite{spathis2022looking} on the weekdays-admitted population to match the size of the weekend one. In the following, we focus on the performance for patients admitted on weekends; we present the converse analysis on the weekdays-admitted patient set and further ablation studies on increasing numbers of outcomes modelled, demonstrating the transportability of the proposed architecture in Appendix~\ref{dj:app:results}.

\paragraph{Evaluation.} We computed survival prediction at the last observation $l_i$ in the 24-hour post-admission period. Models were compared using a time-dependent C-index and Brier score evaluated at time horizons 7 and 30 days after the observation period. Additionally, we adopted an Integrated C-Index to quantify the ranking quality of patients' risks, as risk at a given time horizon in the {ICU} setting may be less relevant than patients' prioritisation. Standard deviations were obtained using 100 bootstrapped iterations on the test set predicted values.


\subsection{Predictive performance}
\begin{figure}[ht]
	\centering
	\includegraphics[width=0.8\textwidth]{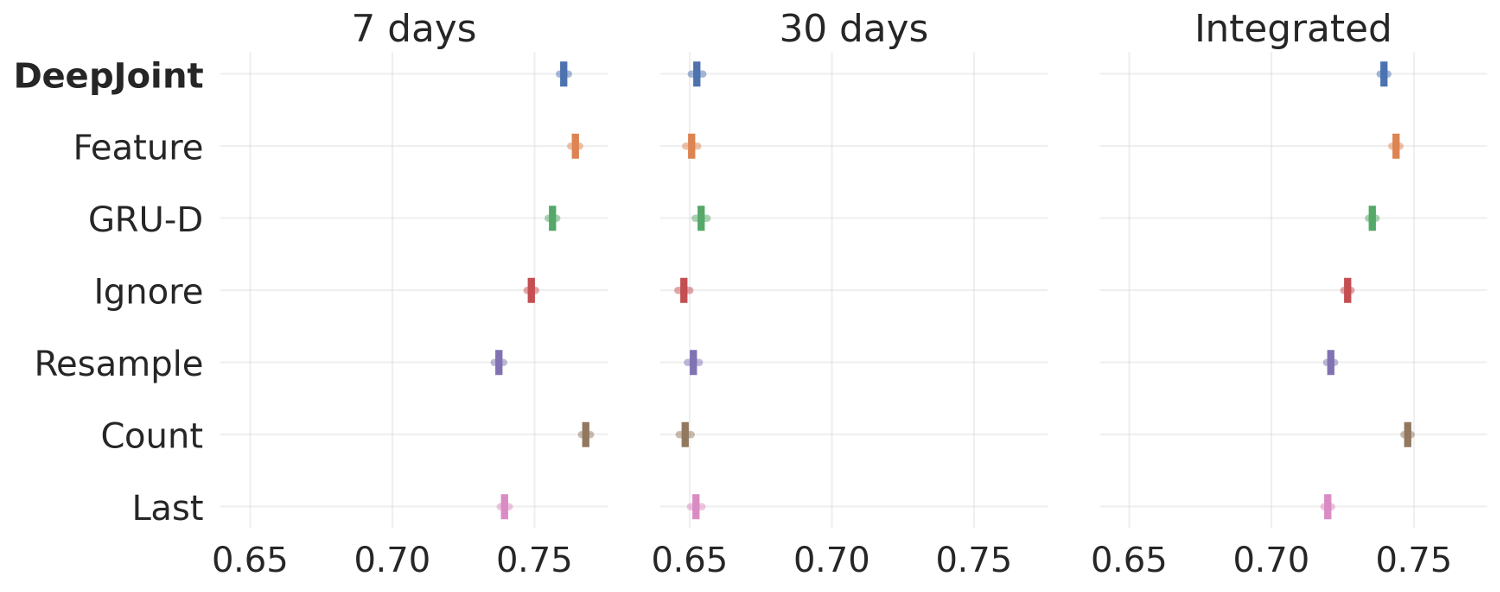}
	\caption{Performance comparison through C-Index on a random split of the \textsc{Mimic III} population with 95\% {CI}. The proposed Deepjoint (in bold) performs similarly to Feature, which uses the same inputs. All methodologies considering clinical presence present improved performance compared to their associated alternatives discarding it.}
	\label{fig:random}
\end{figure}

Figure~\ref{fig:random} describes the models' discriminative performance at different evaluation horizons on the test set, where weekend/weekday status is ignored (see Appendix~\ref{dj:app:results} for Brier score and tabular results). Note that all models rely on a DeepSurv network to model the survival outcome and only differ in their inputs and how they handle irregular timings. 

\paragraph{Insight 1: Integrating clinical presence features improves model performance.} Integrating clinical presence features improves model performance: Count improves over Last, and Feature over Ignore in each of the three considered metrics. Note that these models differ in the input information with Count using the laboratory test count in addition to their last observed values, and Feature including the missingness indicator and time since the last observation in addition to the longitudinal tests modelled by Ignore. The central observation from the integrated metric is that modelling irregular timings improves performance, as shown by the increased performance of {GRU}-D, Feature and DeepJoint over Ignore. Focusing on the proposed Deepjoint, its performances are comparable with Feature, showing that the shared embedding regularisation does not significantly reduce performance. Finally, the simple Count baseline presents one of the best performances in this setting, echoing~\cite{nestor2019feature}'s remark on the superiority of medically relevant features over more complex modelling for improved predictive performances. In the following, we show that these features may be more sensitive to the deployment setting than the proposed methodology.

This first experiment demonstrates the informativeness of irregular timings in modelling survival. Analyses in the literature often stop at this stage. In the following, we propose to study how an internal shift may impact these strategies.

\subsection{Transportability}
\label{mimic:robustness}

\begin{figure}[ht]
	\centering
	\includegraphics[width=\textwidth]{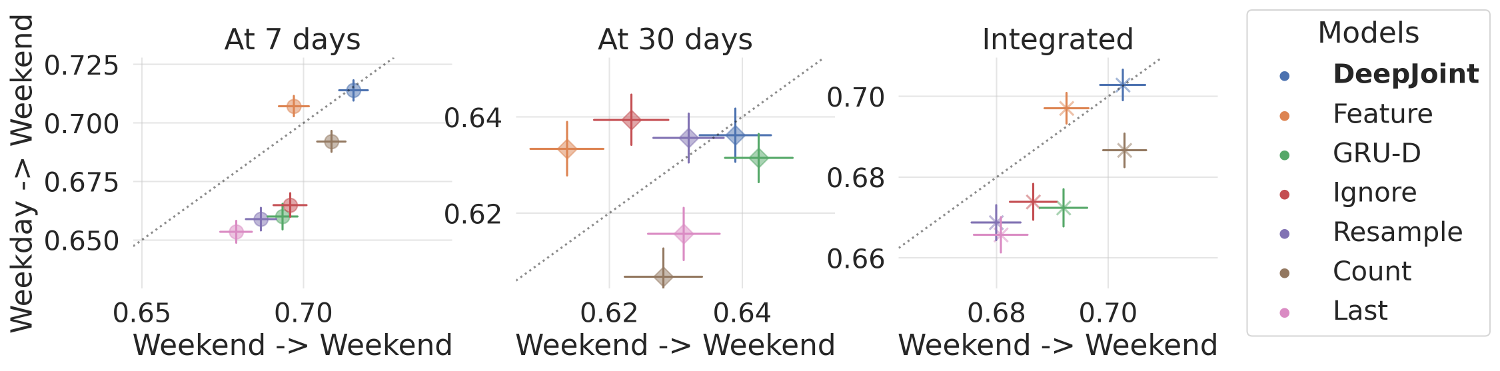}
	\caption{Discriminative performance evaluated on patients admitted on weekends for a transferred model (\textit{y-axis}) and a model trained on weekends-admitted patients and tested on the test set of this same group (\textit{x-axis}) (with 95\% {CI}). Models closer to the diagonal are more transportable. {DeepJoint} presents improved transportability under the weekend effect shift.}
	\label{fig:split_weekend}
\end{figure}

Figure~\ref{fig:split_weekend} presents the performance when a model is transferred from the weekday to the weekend setting (y-axis), and when a model is trained and tested in the same setting (x-axis). A model transportable under shift performs similarly when transferred from another setting as when trained under the same setting, i.e.~models close to the diagonal. This diagonal delimits underfitting (above) from overfitting (under) to the training observational patterns. Practitioners should, therefore, select a model close to the diagonal with the best discriminative performance (upper right corner). As a measure of transportability, we introduce the absolute difference in C-index performance between the model trained and tested in the same setting and the transferred one, referred to as \textit{transfer loss}. The smaller the transfer loss, the more transportable the model is.  Table~\ref{tab:transportability} describes each model's transfer loss at the three considered time horizons used for evaluation.

\paragraph{Insight 2: Modelling approaches that ignore clinical presence are not robust under changes in clinical practice.} In this population, performance differs from the previous random split, as shown on the x-axis. Specifically, laboratory test counts are less informative, resulting in lower performance. However, leveraging irregular timings improves performance, with DeepJoint, Count and Feature presenting the best integrated C-Index. Conversely, models based on the last observation and using resampled data show the worst discriminative performance in this population, aligning with the intuition that ignoring the observational patterns reduces performance. Focusing on the models' transportability, methods ignoring irregular timings or using simple features tend to underperform under shifts. This counterintuitive observation underlines that ignoring irregular timings does not improve transportability. Notably, we would like to echo the remark on the risk of using informative observational patterns raised by~\cite{lipton2016directly, van2020cautionary} in which the authors underline how taking advantage of observational patterns, such as missingness, might lead to a mismatch between the development and deployment settings, but that ignoring this information might not be possible. Finally, note that existing strategies considering irregular timings also drift away from the diagonal, demonstrating their sensitivity to shift. Consequently, existing approaches for modelling irregular timings are not enough to improve transportability. As proposed in this paper, one must model clinical presence as an outcome of the model to extract embeddings that are more transportable under shifts in clinical presence.

\begin{table}[!ht]
	\centering
	\begin{tabular}{r|ccc}
		{} & \multicolumn{3}{c}{Evaluated on weekends} \\
		Model &     7 days  &             30 days  &  Overall\\ \midrule
		\textbf{DeepJoint} & \textbf{0.014} (0.012) & \textbf{0.013} (0.010) & \textbf{0.011} (0.009)\\
		Feature &  \textit{0.015} (0.011) & 0.021 (0.012) & \textbf{0.011} (0.009)\\
		GRU-D & 0.033 (0.015) & \textit{0.015} (0.010) & 0.020 (0.012) \\
		Ignore & 0.032 (0.017) & 0.020 (0.014) & 0.016 (0.012)\\
		Resample &  0.029 (0.015)  & \textit{0.015} (0.010) & \textit{0.015} (0.011)\\
		Count & 0.020 (0.013) & 0.022 (0.012) & 0.017 (0.010)\\
		Last & 0.026 (0.012) & 0.017 (0.012) & \textit{0.015} (0.009) \\
	\end{tabular}
	\caption{Transfer loss - Mean (std). The smaller the loss, the more transportable the model is. Best performances are in \textbf{bold}, second best in \textit{italics}. DeepJoint presents the best transportability properties.}
	\label{tab:transportability}
\end{table}

\paragraph{Insight 3: Joint model improves transportability.} DeepJoint achieves state-of-the-art discrimination performance in both settings as shown in Figure~\ref{fig:split_weekend}. Particularly, when compared with Feature, which uses the same input, our proposed model presents improved performance and transportability with lower transfer loss. This underlines how modelling \emph{jointly} survival and clinical presence regularises the embedding against shifts.

This experiment demonstrates that traditional strategies for handling irregular timings appear to improve performance in internal validation but are prone to overfitting when transferred to a different clinical presence setting. The proposed joint model tackles this challenge under the weekend effect shift. Connecting these results with the previous performance on the random split, a trade-off exists between same setting performance and performance under shift, as hypothesised in~\cite{subbaswamy2020development}. In our experiments, models such as Count and Feature perform best in the random split setting but suffer under clinical presence shifts.

\section{Discussion}
\label{sec:discussion}

In this work, we formalise the problem of clinical presence shift and explore how ML survival models perform under such shifts. Our work demonstrates the shortcomings of existing strategies for handling clinical presence and introduces a novel joint model leveraging neural networks to improve transportability. In this section, we detail our contributions, recommendations, and future works.

\subsection{Contributions}
Several studies have investigated and developed mitigation strategies for a subset of distribution shifts that do not fully capture the complexity of the shifts observed in medical data. In this work, we distinguish \textit{observed} covariates from \textit{observable} ones. This simple distinction highlights a new type of shift: \textit{clinical presence shift}. This formalisation offers a new lens through which to understand distribution shifts in medical data. Critically, the underlying distributions may remain stable, but the observation process may evolve, resulting in a shift in observed distributions. This contribution has important repercussions on the development of clinical prediction models, as their adoption in medical practice can alter the clinical presence process.

The central challenge with clinical presence shifts is that one would like to take advantage of its informativeness while ensuring the model's transportability under an evolving process. At the intersection of adversarial {ML} and statistical joint model, we introduce a multitask neural network that models the survival outcome and the irregular timings associated with clinical presence, namely temporality and missingness. The motivation behind this joint modelling is the mounting evidence of improved performance and theoretical robustness when considering multiple tasks. Our analysis of the \textsc{Mimic III} dataset demonstrates that the practice of ignoring clinical presence lowers predictive performances. Not only does it reduce performance in a given setting, but counter-intuitively, ignoring the observation process reduces transportability under shift. On the contrary, the proposed method presents improved transportability properties.



\subsection{Recommendations}
We demonstrate the critical importance of considering clinical presence for improving prediction models' performance and transportability. Specifically, we invite model developers to:

\begin{enumerate} 
	\item \textit{Model clinical presence.} Our work introduces a novel way to not only featurise clinical presence but also model it. As shown in this work, this approach presents a step towards transportability. 
	\item \textit{Leverage natural experiments.} While the necessity of transportability across hospitals is sometimes questioned~\cite{futoma2020myth}, we argue that models should be transportable under naturally occurring shifts \textit{in a given hospital}. We invite model developers to take advantage of natural experiments to study such transportability properties, as proposed in the weekend effect analysis in Section~\ref{deepjoint:sec:experiment}.
\end{enumerate}

\subsection{Future work}
Our analysis empirically evidences the importance of modelling clinical presence, which opens multiple avenues for future work. First, the proposed formalisation invites further theoretical work and alternative mitigation strategies as multitasking remains empirically superior, but the theory still needs to be developed. Specifically, while the method empirically improves transportability in the studied setting, a theoretical bound on its robustness to clinical presence shifts would better support safe deployment. Then, on the empirical side, to confirm the results presented in this work, we will explore additional natural experiments, such as the July effect~\cite{young2011july}, or grouping practitioners by expertise, and examine other datasets beyond \textsc{Mimic-III}, which may be more comprehensive than commonly collected EHRs and allow inter-hospital transportability analysis. Finally, our theoretical analysis demonstrates transportability under small clinical presence shifts. Our future work will explore shifts under which the proposed approach fails, and evaluate the model's capacity to \textit{detect} such shifts.

\begin{acknowledgements}
	This work was supported by the UK Medical Research Council programme (grant MC\_UU\_00002/5 and MC\_UU\_00002/2) and theme (grant MC\_UU\_00040/02 – Precision Medicine) funding. For the purpose of open access, the author has applied a Creative Commons Attribution (CC BY) licence to any Author Accepted Manuscript version arising from this submission
\end{acknowledgements}

%
%

\bibliographystyle{abbrv}
\bibliography{bibliography}   

\newpage
\begin{appendix}
	\section{Mimic III - Experiments}
	\label{app:experiments}
	
	\subsection{Data characteristics}
\begin{figure}[!ht]
		\centering
		\includegraphics[width=0.4\textwidth]{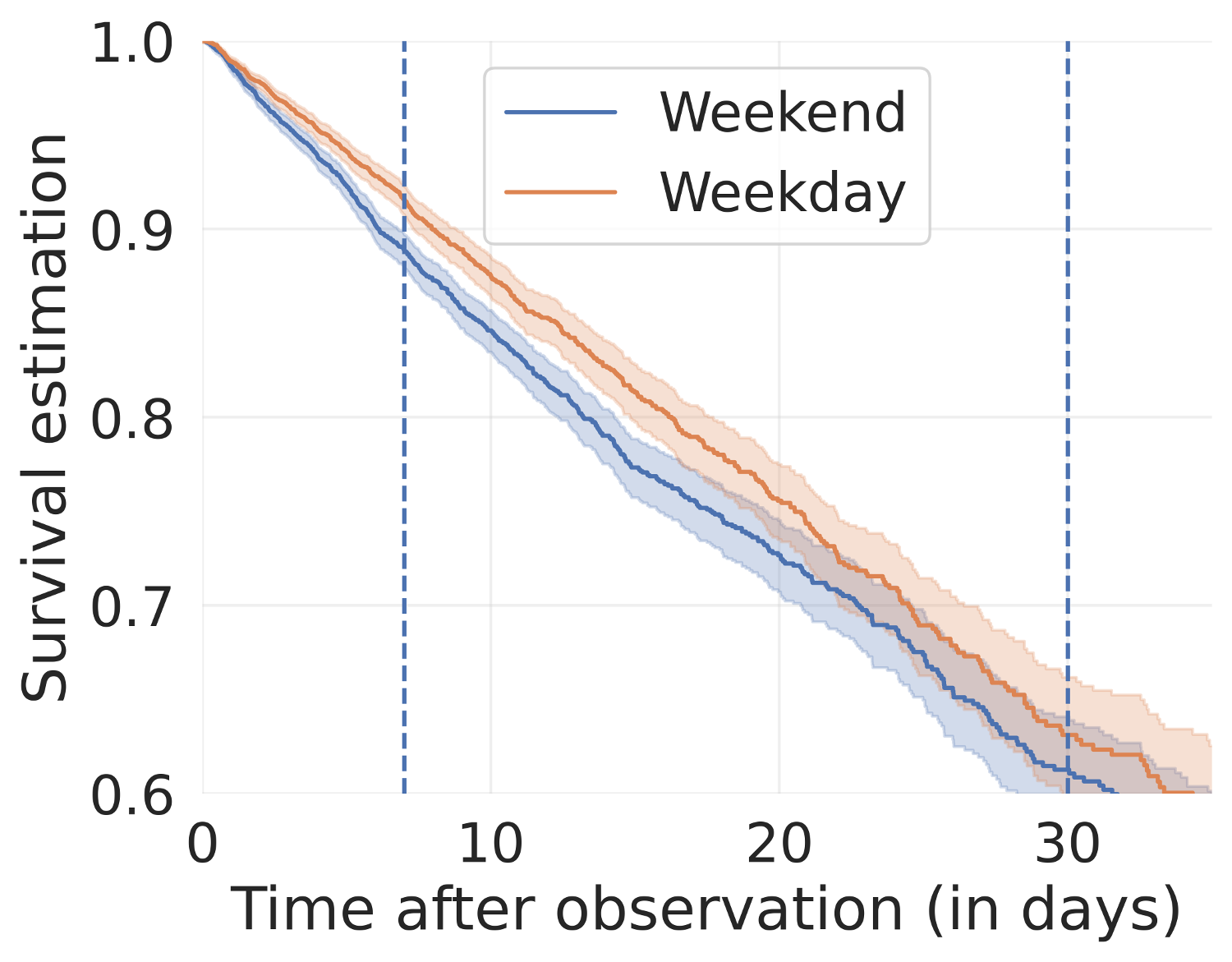}
		\caption{Kaplan-Meier survival estimates following 24 hours of observation stratified by the admission group. Dotted lines denote evaluation horizons.}
		\label{fig:survival_mimic}
	\end{figure}

	Table~\ref{tab:mimic:pop} presents the demographic characteristics of the studied population, and Table~\ref{tab:mimic:labs} summarises the set of tests selected with the mean number of tests performed during the 24 hours post-admission and their mean values. The results are presented at the population level and differentiated by the subgroups used to study the impact of clinical presence shift. Finally, Figure~\ref{fig:survival_mimic} displays the Kaplan-Meier survival estimates following 24 hours of observation.
	\begin{table}[!ht]
		\centering
		\small
		\begin{threeparttable}
			\begin{tabular}{ccr|ccc}
				&&& \multicolumn{3}{c}{Population}\\\cmidrule{4-6}
				&    &      &  Overall &  Weekday &  Weekend \\
				\midrule
				\multicolumn{3}{r|}{Number of patients} &31,692& 24,865& 6,827\\
				\midrule
				\multirow{5}{*}{Outcome} & \multicolumn{2}{r|}{Length of stay (in days$^*$)} & 10.00 & 9.97 & 10.11\\\cmidrule{3-6}
				&\multirow{4}{*}{Death} &Overall (\%)  &  13.23 & 12.69 & 15.19 \\
				&& 7 days$^+$ (\%) & 6.90& 6.45& 8.48\\
				&& 30 days$^+$ (\%)  &  12.28& 11.73& 14.30\\
				\midrule
				\multirow{10}{*}{Demographics} 
				&\multirow{2}{*}{Gender} & Male (\%) &  57.12 & 57.36 & 56.22 \\
				&& Female (\%)&  42.88 & 42.64 & 43.78 \\
				\cmidrule{3-6}
				&\multirow{5}{*}{Ethnicity} & Asian (\%)&  2.34 & 2.34 & 2.34   \\
				&& Black (\%)&    7.39 & 7.28 & 7.78 \\
				& & Hispanic (\%)&   3.16 & 3.06 & 3.52\\
				& & White (\%)&    71.34 & 72.02 & 68.87  \\
				& & Other (\%) &   15.77 & 15.30 & 17.49\\
				\cmidrule{3-6}
				& \multirow{3}{*}{Insurance} & Public (\%)&   65.42 & 65.07 & 66.69  \\
				& & Private (\%) &    33.45 & 33.97 & 31.55\\
				& & Self Pay (\%)&   1.13 & 0.96 & 1.76 \\
			\end{tabular}
			\begin{tablenotes}
				\small
				\item $^*$ Mean (std)
				\item $^+$ After first day of observation
			\end{tablenotes}
			\caption{\textsc{Mimic III} - Population characteristics between patients admitted on weekdays and weekends.}
			\label{tab:mimic:pop}
		\end{threeparttable}
	\end{table}
	
	\begin{table}[!ht]
		\small
		\addtolength{\tabcolsep}{-2pt}
		\begin{threeparttable}
			\begin{tabular}{r|ccc||ccc}
				&  \multicolumn{3}{c||}{Lab Value}&  \multicolumn{3}{c}{Number Test}\\
				Laboratory test & Weekday &  &Weekend & Weekday &  &Weekend \\
				\midrule
				Anion gap & 13.73 (3.24) &  * &14.15 (3.28) & 1.78 (1.03) &  * &1.92 (1.09) \\
				Bicarbonate & 24.17 (4.23) &  * &23.79 (4.56) & 1.87 (1.00) &  * &1.94 (1.09) \\
				Blood urea nitrogen & 24.51 (19.83) &  * &25.96 (21.50) & 1.90 (0.99) &  * &1.95 (1.08) \\
				Calcium (total) & 8.37 (0.80) &  * &8.31 (0.77) & 1.45 (1.10) &  * &1.69 (1.10) \\
				Chloride & 105.30 (5.70) &   &105.25 (5.98) & 1.92 (1.03) &  * &1.98 (1.12) \\
				Creatinine & 1.30 (1.34) &  * &1.36 (1.36) & 1.91 (0.99) &  * &1.96 (1.08) \\
				Glucose & 137.85 (48.08) &   &137.75 (53.83) & 3.32 (2.99) &  * &2.46 (2.00) \\
				Hematocrit & 32.63 (5.37) &  * &32.88 (5.39) & 3.30 (2.62) &  * &2.66 (2.03) \\
				Hemoglobin & 11.00 (1.92) &  * &11.10 (1.93) & 2.73 (2.31) &  * &2.09 (1.71) \\
				Magnesium & 1.99 (0.39) &  * &1.97 (0.34) & 1.68 (1.07) &  * &1.85 (1.10) \\
				MCH$^1$ & 30.31 (2.48) &   &30.30 (2.58) & 1.79 (1.01) &  * &1.72 (1.01) \\
				MCH$^1$ conc. & 33.90 (1.51) &  * &33.80 (1.56) & 1.79 (1.01) &  * &1.72 (1.01) \\
				MCV$^2$ & 89.49 (6.54) &  * &89.71 (6.77) & 1.79 (1.01) &  * &1.72 (1.01) \\
				PTH$^3$ & 38.02 (20.04) &  * &38.94 (21.37) & 1.55 (1.30) &   &1.56 (1.37) \\
				Phosphate & 3.59 (1.16) &  * &3.49 (1.17) & 1.44 (1.11) &  * &1.69 (1.10) \\
				Platelets & 219.89 (106.52) &  * &224.16 (114.58) & 1.94 (1.13) &  * &1.83 (1.15) \\
				Potassium & 4.16 (0.55) &  * &4.10 (0.56) & 1.98 (1.10) &  * &2.11 (1.16) \\
				Red blood cell & 5.26 (15.58) &   &5.66 (18.21) & 1.91 (1.06) &  * &1.84 (1.08) \\
				Sodium & 138.84 (4.33) &  * &139.03 (4.69) & 1.88 (1.09) &  * &2.01 (1.16) \\
				White blood cell & 11.98 (13.90) &   &12.18 (15.34) & 1.92 (1.06) &  * &1.86 (1.09) \\
				pH & 6.93 (0.76) &  * &6.85 (0.80) & 3.16 (3.74) &  * &2.30 (2.94) \\
			\end{tabular}
			\begin{tablenotes}
				\small
				\item $^1$ Mean corpuscular hemoglobin.
				\item $^2$ Mean corpuscular volume.
				\item $^3$ Partial thromboplastin time.
				\item * signifies that a 2-sided T-test has a p-value < 0.05
			\end{tablenotes}
			\caption{List of laboratory tests used with the associated mean number of tests and values (and standard deviations) differentiated by admission populations.}
			\label{tab:mimic:labs}
		\end{threeparttable}
	\end{table}
	
	\clearpage
	\subsection{Hyperparameters tuning}
	All models' hyper-parameters were selected over the following grid of hyperparameters (if appropriate) following 100 iterations of random search. All experiments were ran on a A100 GPU over 100 hours for each experiment.
	
	\begin{table}[!ht]
		\centering
		\small
		\begin{threeparttable}
			\begin{tabular}{cr|l}
				& Hyperparameter        & Values                   \\
				\midrule
				\multirow{4}{*}{Training}                  & Learning rate         & $10^{-3}$, $10^{-4}$ \\
				& Batch size            & $512$ , $1024$\\
				& $\alpha$              & $0.1$, $0.3$\\
				& $\theta$                     & $2^*$                     \\
				\midrule
				\multirow{2}{*}{RNN}                       & Layers                & $1$, $2$\\
				& Hidden nodes          & $10$, $25$                                      \\
				\midrule
				\multirow{2}{*}{Survival}                  & Layers                & $0$, $1$, $2$, $3$                    \\
				& Nodes                 & $50$      \\
				\midrule
				\multirow{2}{*}{Clinical Presence}
				& Temporal ($I$)             & Same parameters explored as survival           \\
				& Missingnesss ($M$)           & Same parameters explored as survival           \\
			\end{tabular}
			\begin{tablenotes}
				\small
				\item $^*$ Following the results from~\cite{liu2019end}.
			\end{tablenotes}
			\caption{Grid used for hyperparameters search}
			\label{tab:grid}
		\end{threeparttable}
	\end{table}
	
	\subsection{Modelling clinical presence}
	\label{dj:app:results}
	
	Table~\ref{tab:res} presents the discriminative results on the test set of the first set of experiments for which patients are randomly split into train and test sets. 
	\begin{table}[!ht]
		\small
		\addtolength{\tabcolsep}{-2pt}
		\centerline{
			\begin{tabular}{r|ccc||ccc}
				\multirow{2}{*}{Model} & \multicolumn{3}{c||}{C Index} & \multicolumn{3}{c}{Brier Score} \\
				&             7  &             30  &             Overall  &            7  &             30  &             Overall \\\midrule
				\textbf{DeepJoint}  & 0.760 (0.008) & \textit{0.652} (0.010) & 0.739 (0.007) & 0.076 (0.002) & \textit{0.267} (0.011) & \textbf{0.127} (0.084) \\
				Feature                  &  \textit{0.764} (0.008) & 0.651 (0.011) & \textit{0.744} (0.007)  & \textit{0.075} (0.002) & 0.272 (0.012) & - \\
				GRU-D                    & 0.756 (0.008) & \textbf{0.654} (0.011) & 0.735 (0.007) &  \textbf{0.074} (0.002) & \textbf{0.266} (0.011) & 0.160 (0.049)\\
				Resample                 & 0.737 (0.009) & 0.651 (0.011) & 0.721 (0.008) &  0.076 (0.002) & 0.270 (0.011) & - \\
				Ignore                   & 0.749 (0.008) & 0.648 (0.011) & 0.727 (0.007) &  \textit{0.075} (0.002) & 0.273 (0.011) & -\\
				Count                    & \textbf{0.768} (0.008) & 0.648 (0.011) & \textbf{0.748} (0.006) &  \textit{0.075} (0.002) & 0.278 (0.012) & \textit{0.144} (0.081) \\
				Last                     &  0.739 (0.008) & \textit{0.652} (0.011) & 0.720 (0.007) & \textit{0.075} (0.002) & 0.269 (0.011) & 0.145 (0.066)
		\end{tabular}}
		\caption{Comparison of model performance by means (standard deviations) in the random split experiment of \textsc{Mimic III} dataset. Best performances are in \textbf{bold}, second best in \textit{italics}. '-' denotes the divergence of the Brier score.}
		\label{tab:res}
	\end{table}
	
	\newpage
	\subsection{Transfer performances}
	\label{dj:app:subsample}
	
\begin{figure}[!ht]
		\centering
		\includegraphics[width=\textwidth]{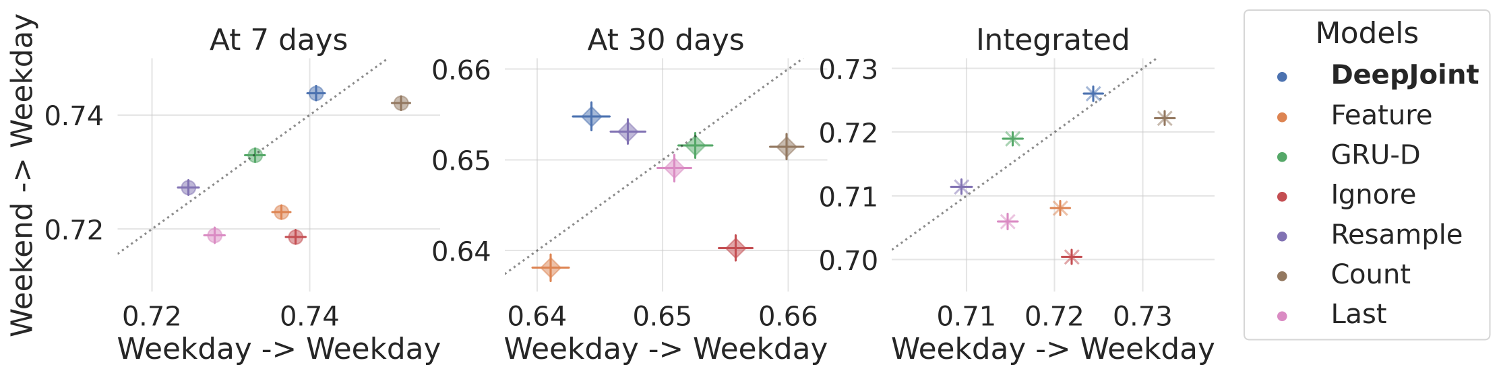}
		\caption{Transportability evaluation for patients admitted on weekdays.}
		\label{fig:split_weekday}
	\end{figure}

	\paragraph{Weekday Performance.} Figure~\ref{fig:split_weekday} shows the performance on patients admitted on weekdays. Performance resembles the one presented in Section~\ref{mimic:robustness} with DeepJoint presenting better transportability than Feature, its alternative that does not model clinical presence as a model's output, except at 30 days where the model underfit. Count presents the best internal performance in this setting but suffers when transferred as counts during weekends and weekdays may reflect different information. An important observation is that the proposed method does not overfit in the presented setting, as the multitask learning regularises the shared embedding.
	
	\paragraph{Tabular Performance.} Tables~\ref{tab:transfer7},~\ref{tab:transfer30} and~\ref{tab:transferall} present the discriminative difference on the weekends-admissions test set between the model trained on weekends-admissions and the model transferred from weekdays to weekends (and the opposite scenario). 
	
	These results emphasise that the proposed joint modelling \textbf{DeepJoint} is more transportable than state-of-the-art approaches based on the same inputs. Interestingly, ignoring the clinical process (\textbf{Resample} or \textbf{Ignore}) seems to be less transportable under shifts in the observation process, echoing the remark made by~\cite{lipton2016directly} about the difficulty of ignoring it. Finally, note how simple features such as in \textbf{Count} might be detrimental under shift --- particularly pronounced when applied to weekends-admissions.
	
	\begin{table}[!ht]
		\small
		\addtolength{\tabcolsep}{-2pt}
		\centerline{
			\begin{tabular}{r|ccc||ccc}
				{} & \multicolumn{6}{c}{Horizon: 7 days after observation}\\
				{} & \multicolumn{3}{c}{Evaluated on weekends}& \multicolumn{3}{c}{Evaluated on weekdays}\\\cmidrule{2-4}\cmidrule{5-7}
				Model &             Transfer  &             Internal  &  Difference &                    Transfer  &             Internal  &  Difference  \\\midrule
				\textbf{DeepJoint} & \textbf{0.714} (0.022) & \textbf{0.715} (0.023) & \textbf{0.014} (0.012) & \textbf{0.744} (0.006) & \textit{0.741} (0.006) & \textit{0.005} (0.003) \\
				Feature & \textit{0.707} (0.022) & 0.697 (0.024) & \textit{0.015} (0.011) & 0.722 (0.006) & 0.737 (0.006) & 0.014 (0.005) \\
				GRU-D & 0.660 (0.028) & 0.693 (0.024) & 0.033 (0.015)& 0.733 (0.006) & 0.732 (0.007) & \textbf{0.003} (0.002) \\
				Ignore & 0.665 (0.026) & 0.696 (0.026) & 0.032 (0.017) & 0.718 (0.006) & 0.738 (0.007) & 0.019 (0.005) \\
				Resample & 0.659 (0.025) & 0.687 (0.024) & 0.029 (0.015) & 0.727 (0.007) & 0.724 (0.007) & \textit{0.005} (0.003) \\
				Count & 0.692 (0.023) & \textit{0.709} (0.022) & 0.020 (0.013) & \textit{0.742} (0.006) & \textbf{0.751} (0.006) & 0.010 (0.004) \\
				Last & 0.653 (0.024) & 0.679 (0.025) & 0.026 (0.012) & 0.718 (0.007) & 0.727 (0.007) & 0.009 (0.004) \\
		\end{tabular}}
		\caption{C-Index performance at 7 days for models trained and tested on the same type of patients (Internal), transferred from the other setting (Transfer) and their difference after the first day of observation - Mean (std). Best performances are in \textbf{bold}, second best in \textit{italics}. }
		\label{tab:transfer7}
	\end{table}
	
	\begin{table}[!ht]
		\small
		\addtolength{\tabcolsep}{-2pt}
		\centerline{
			\begin{tabular}{r|ccc||ccc}
				{} & \multicolumn{6}{c}{Horizon: 30 days after observation}\\
				{} & \multicolumn{3}{c}{Evaluated on weekends}& \multicolumn{3}{c}{Evaluated on weekdays}\\\cmidrule{2-4}\cmidrule{5-7}
				Model &             Transfer  &             Internal  &  Difference &                    Transfer  &             Internal  &  Difference  \\\midrule
				\textbf{DeepJoint} & \textit{0.636} (0.028) & \textit{0.639} (0.027) & \textbf{0.013} (0.010) & \textbf{0.655} (0.008) & 0.644 (0.007) & 0.011 (0.005) \\
				Feature & 0.633 (0.028) & 0.614 (0.028) & 0.021 (0.012) & 0.638 (0.007) & 0.641 (0.007) & \textit{0.004} (0.004) \\
				GRU-D & 0.631 (0.026) & \textbf{0.642} (0.026) & \textit{0.015} (0.010) & 0.652 (0.007) & 0.653 (0.007) & \textbf{0.003} (0.002) \\
				Ignore & \textbf{0.639} (0.027) & 0.623 (0.029) & 0.020 (0.014) & 0.640 (0.007) & \textit{0.656} (0.007) & 0.016 (0.006) \\
				Resample & \textit{0.636} (0.026) & 0.632 (0.027) & \textit{0.015} (0.010) & \textit{0.653} (0.007) & 0.647 (0.007) & 0.007 (0.004) \\
				Count & 0.607 (0.030) & 0.628 (0.030) & 0.022 (0.012) & 0.651 (0.007) & \textbf{0.660} (0.007) & 0.009 (0.005) \\
				Last & 0.616 (0.028) & 0.631 (0.028) & 0.017 (0.012) & 0.649 (0.007) & 0.651 (0.007) & \textit{0.004} (0.003) \\
		\end{tabular}}
		\caption{C-Index performance at 30 days for models trained and tested on the same type of patients (Internal), transferred from the other setting (Transfer) and their difference after the first day of observation - Mean (std). Best performances are in \textbf{bold}, second best in \textit{italics}.}
		\label{tab:transfer30}
	\end{table}
	
	\begin{table}[!ht]
		\small
		\addtolength{\tabcolsep}{-2pt}
		\centerline{
			\begin{tabular}{r|ccc||ccc}
				{} & \multicolumn{6}{c}{Integrated}\\
				{} & \multicolumn{3}{c}{Evaluated on weekends}& \multicolumn{3}{c}{Evaluated on weekdays}\\\cmidrule{2-4}\cmidrule{5-7}
				Model &             Transfer  &             Internal  &  Difference &                    Transfer  &             Internal  &  Difference  \\\midrule
				\textbf{DeepJoint} & \textbf{0.703} (0.019) & \textbf{0.703} (0.020) & \textbf{0.011} (0.009)  & \textbf{0.726} (0.006) & \textit{0.724} (0.006) & \textbf{0.004} (0.003) \\
				Feature & \textit{0.697} (0.019) & \textit{0.693} (0.020) & \textbf{0.011} (0.009)  & 0.708 (0.006) & 0.721 (0.006) & 0.013 (0.004) \\
				GRU-D & 0.672 (0.024) & 0.692 (0.022) & 0.020 (0.012) & 0.719 (0.005) & 0.715 (0.006) & \textbf{0.004} (0.003) \\
				Ignore & 0.674 (0.023) & 0.687 (0.021) & 0.016 (0.012) & 0.700 (0.006) & 0.722 (0.006) & 0.021 (0.005) \\
				Resample & 0.669 (0.022) & 0.680 (0.022) & \textit{0.015} (0.011) & 0.711 (0.006) & 0.709 (0.006) & \textbf{0.004} (0.003) \\
				Count & 0.687 (0.021) & \textbf{0.703} (0.020) & 0.017 (0.010) & \textit{0.722} (0.006) & \textbf{0.732} (0.006) & 0.010 (0.004) \\
				Last & 0.666 (0.022) & 0.681 (0.024) & \textit{0.015} (0.009) & 0.706 (0.006) & 0.715 (0.006) & \textit{0.009} (0.004) \\
		\end{tabular}}
		\caption{Integrated C-Index performance for models trained and tested on the same type of patients (Internal), transferred from the other setting (Transfer) and their difference after the first day of observation - Mean (std).  Best performances are in \textbf{bold}, second best in \textit{italics}.}
		\label{tab:transferall}
	\end{table}

	\newpage
	\section{Ablation studies}
	
	

	\subsection{Impact of \textit{I} and \textit{M} networks on performance}
	In this section, we explore how the different dimensions of clinical presence impact performance. Patients were randomly assigned to train and test sets as presented in the first set of experiments. Based on the laboratory values, missingness indicator and time of observations as inputs of the network, we compared all possible combinations of components as Table~\ref{table:summary:model} summarises. 
	
	\begin{table}[!ht]
		\centering
		\small
		\begin{tabular}{r|ccc}
			Model      & $S$ & $I$ & $M$\\\midrule
			\textbf{DeepJoint}& $\bullet$     &  $\bullet$ & $\bullet$ \\
			DeepJoint$_M$  & $\bullet$     &            & $\bullet$\\
			DeepJoint$_I$  & $\bullet$       & $\bullet$  &           \\
			Feature      & $\bullet$       &            &          \\
		\end{tabular}
		\caption{Components associated to each architecture.}
		\label{table:summary:model}
	\end{table}
	
	Table~\ref{tab:mimic:ablation} presents the discriminative performance of these different models. Note how each individual model performs similarly to \textbf{Feature} baseline based on the same inputs. 
	
	\begin{table}[!ht]
		\small
		\addtolength{\tabcolsep}{-2pt}
		\centerline{
			\begin{tabular}{r|ccc||ccc}
				\multirow{2}{*}{Model} & \multicolumn{3}{c||}{C Index} & \multicolumn{3}{c}{Brier Score} \\
				&             7&             30&             Integrated&             7&             30&             Integrated\\\midrule
				\textbf{DeepJoint}  &                      0.760 (0.008) &\textbf{0.652} (0.010) &0.739 (0.007) &  \textit{0.076} (0.002) &\textbf{0.267} (0.011) & \textbf{0.127} (0.084)\\
				DeepJoint$_M$                   & \textbf{0.764} (0.008) & \textbf{0.652} (0.011) & \textbf{0.744 }(0.007) &   \textbf{0.075} (0.002) & \textbf{0.267} (0.011) & 0.145 (0.068) \\
				DeepJoint$_I$                    &\textit{0.762} (0.008) & 0.650 (0.011) & \textit{0.741} (0.007) &  \textit{0.076} (0.002) & 0.279 (0.013) & - \\
				Feature                  &        \textbf{0.764} (0.008) & \textit{0.651} (0.011) & \textbf{0.744} (0.007)  & \textbf{0.075} (0.002) & 0.272 (0.012) & - \\
				
		\end{tabular}}
		\caption{Comparison of model performance by means (standard deviations) in the random split experiment of \textsc{Mimic III} dataset. Best performances are in \textbf{bold}, second best in \textit{italics}.}
		\label{tab:mimic:ablation}
	\end{table}

	\subsection{Impact of \textit{I} and \textit{M} networks on transportability}
	From Theorem~\eqref{theorem}, more uncorrelated tasks are modelled, more robust would be the proposed architecture. Therefore, we investigated the transportability of the previous models. The smaller difference between settings shows that DeepJoint$_I$ and DeepJoint$_M$ improve above the Feature baseline, but their combination outperforms each individual.

	\begin{table}[!ht]
		\small
		\addtolength{\tabcolsep}{-2pt}
		\centerline{
			\begin{tabular}{r|ccc||ccc}
				{} & \multicolumn{6}{c}{Horizon: 7 days after observation}\\
				{} & \multicolumn{3}{c}{Evaluated on weekends}& \multicolumn{3}{c}{Evaluated on weekdays}\\\cmidrule{2-4}\cmidrule{5-7}
				Model &             Transfer  &             Internal  &  Difference &                    Transfer  &             Internal  &  Difference  \\\midrule
				\textbf{DeepJoint} & \textit{0.714} (0.022) & \textbf{0.715} (0.023) & \textbf{0.014} (0.012) & \textbf{0.744} (0.006) & \textit{0.741} (0.006) & \textbf{0.005} (0.003) \\
				DeepJoint$_M$ & \textbf{0.715} (0.023) & \textit{0.713} (0.026) & 0.019 (0.014) & 0.732 (0.007) & \textbf{0.743} (0.006) & 0.011 (0.006)\\
				DeepJoint$_I$ & 0.705 (0.022) & \textbf{0.715} (0.022) & 0.018 (0.013) & \textit{0.739} (0.006) & 0.735 (0.006) & \textit{0.006} (0.004)\\
				Feature & 0.707 (0.022) & 0.697 (0.024) & \textit{0.015} (0.011) & 0.722 (0.006) & 0.737 (0.006) & 0.014 (0.005) \\
				
		\end{tabular}}
		\caption{C-Index performance at 7 days for models trained and tested on the same type of patients (Internal), transferred from the other setting (Transfer) and their difference after the first day of observation - Mean (std). Best performances are in \textbf{bold}, second best in \textit{italics}.}
		\label{tab:abl:transfer7}
	\end{table}
	
	\begin{table}[!ht]
		\small
		\addtolength{\tabcolsep}{-2pt}
		\centerline{
			\begin{tabular}{r|ccc||ccc}
				{} & \multicolumn{6}{c}{Horizon: 30 days after observation}\\
				{} & \multicolumn{3}{c}{Evaluated on weekends}& \multicolumn{3}{c}{Evaluated on weekdays}\\\cmidrule{2-4}\cmidrule{5-7}
				Model &             Transfer  &             Internal  &  Difference &                    Transfer  &             Internal  &  Difference  \\\midrule
				\textbf{DeepJoint} & \textit{0.636} (0.028) & \textit{0.639} (0.027) & \textbf{0.013} (0.010) & \textbf{0.655} (0.008) & \textit{0.644} (0.007) & 0.011 (0.005) \\
				DeepJoint$_M$ & \textbf{0.648} (0.028) & 0.637 (0.027) & \textit{0.018} (0.013) & \textit{0.648} (0.008) & \textbf{0.648} (0.007) & \textit{0.005} (0.004) \\
				DeepJoint$_I$ & 0.631 (0.029) & \textbf{0.647} (0.028) & 0.019 (0.013) & \textit{0.648} (0.008) & 0.639 (0.007) & 0.009 (0.006) \\
				Feature & 0.633 (0.028) & 0.614 (0.028) & 0.021 (0.012) & 0.638 (0.007) & 0.641 (0.007) & \textbf{0.004} (0.004) \\
		\end{tabular}}
		\caption{C-Index performance at 30 days for models trained and tested on the same type of patients (Internal), transferred from the other setting (Transfer) and their difference after the first day of observation - Mean (std). }
		\label{tab:abl:transfer30}
	\end{table}
	
	\begin{table}[!ht]
		\small
		\addtolength{\tabcolsep}{-2pt}
		\centerline{
			\begin{tabular}{r|ccc||ccc}
				{} & \multicolumn{6}{c}{Integrated}\\
				{} & \multicolumn{3}{c}{Evaluated on weekends}& \multicolumn{3}{c}{Evaluated on weekdays}\\\cmidrule{2-4}\cmidrule{5-7}
				Model &             Transfer  &             Internal  &  Difference &                    Transfer  &             Internal  &  Difference  \\\midrule
				\textbf{DeepJoint} & \textit{0.703} (0.019) & \textit{0.703} (0.020) & \textbf{0.011} (0.009) & \textbf{0.726} (0.006) & \textit{0.724} (0.006) & \textbf{0.004} (0.003) \\
				DeepJoint$_M$ & \textbf{0.706} (0.019) & \textbf{0.705} (0.022) & 0.015 (0.011) & 0.716 (0.006) & \textbf{0.726} (0.006) & \textit{0.010} (0.004)\\
				DeepJoint$_I$ & 0.696 (0.019) & 0.702 (0.020) & 0.013 (0.011) & \textit{0.720} (0.006) & 0.719 (0.006) & \textbf{0.004} (0.003) \\
				Feature & 0.697 (0.019) & 0.693 (0.020) & \textbf{0.011} (0.009) & 0.708 (0.006) & 0.721 (0.006) & 0.013 (0.004)\\
		\end{tabular}}
		\caption{Integrated C-Index performance for models trained and tested on the same type of patients (Internal), transferred from the other setting (Transfer) and their difference after the first day of observation - Mean (std).  Best performances are in \textbf{bold}, second best in \textit{italics}.}
		\label{tab:abl:transferall}
	\end{table}
\end{appendix}

\end{document}